\documentclass{article}

\usepackage{graphicx}
\usepackage[caption=false]{subfig}

\begin{document}

\title{Application of Quantum Pre-Processing Filter for Binary Image Classification with Small Samples}

\maketitle

\author{Farina Riaz, Shahab Abdulla, Hajime Suzuki, Srinjoy Ganguly, Ravinesh C.~Deo, and Susan Hopkins}

%\begin{affiliations}
%F. Riaz, H. Suzuki\\
%Commonwealth Scientific and Industrial Research Organization\\
%Sydney, NSW 2000, Australia\\
%Email Address: farina.riaz@data61.csiro.au
%
%F. Riaz, S. Abdulla, S. Ganguly,  S. Hopkins\\
%UniSQ College, University of Southern Queensland\\
%Toowoomba, QLD 4350, Australia
%
%R. C. Deo\\
%School of Mathematics, Physics and Computing, University of Southern Queensland\\
%Springfield, QLD 4300, Australia
%\end{affiliations}

%\keywords{Quantum Computing, Machine Learning, Image Classification, Neural Network}

% Abstract should be written in the present tense and impersonal style (i.e., avoid we), and be at most 200 words long
\begin{abstract}
Over the past few years, there has been significant interest in Quantum Machine Learning (QML) among researchers, as it has the potential to transform the field of machine learning. Several models that exploit the properties of quantum mechanics have been developed for practical applications. In this study, we investigated the application of our previously proposed quantum pre-processing filter (QPF) to binary image classification. We evaluated the QPF on four datasets: MNIST (handwritten digits), EMNIST (handwritten digits and alphabets), CIFAR-10 (photographic images) and GTSRB (real-life traffic sign images). Similar to our previous multi-class classification results, the application of QPF improved the binary image classification accuracy using neural network against MNIST, EMNIST, and CIFAR-10 from 98.9\%\ to 99.2\%, 97.8\%\ to 98.3\%, and 71.2\%\ to 76.1\%, respectively, but degraded it against GTSRB from 93.5\%\ to 92.0\%. We then applied QPF in cases using a smaller number of training and testing samples, i.e. 80 and 20 samples per class, respectively. In order to derive statistically stable results, we conducted the experiment with 100 trials choosing randomly different training and testing samples and averaging the results. The result showed that the application of QPF did not improve the image classification accuracy against MNIST and EMNIST but improved it against CIFAR-10 and GTSRB from 65.8\%\ to 67.2\%\ and 90.5\%\ to 91.8\%, respectively. Further research will be conducted as part of future work to investigate the potential of QPF to assess the scalability of the proposed approach to larger and complex datasets.
\end{abstract}

\section{Introduction}

Over the past few years, there has been significant interest in Quantum Machine Learning (QML), with various algorithms proposed for image processing \cite{4314}. Quantum machine learning has been a hot topic recently \cite{4507}, especially since quantum hardware development has gradually accelerated \cite{4509}. The application of quantum technology in image processing is crucial for efficiently extracting valuable information from real-world scenarios. Numerous approaches have been developed for quantum image classification, such as quantum neural networks \cite{3905}, quantum convolutional neural network \cite{3888}, hybrid quantum classical convolutional neural network \cite{3923}, quantum generative adversarial network \cite{4402,4406} and quantum support vector machines \cite{3841}. The goal of using QML in images is to extract essential features from the image. To achieve this, a classical kernel approach can first be used to estimate unsolvable quantum kernels on a quantum device. Secondly, different models can be created that process the feature vectors using quantum models based on variational circuits. These models gain their strengths by outsourcing nonlinearity into the process of encoding inputs into a quantum state or the quantum feature map. This combination of quantum computing with kernel theory will help in developing QML algorithms that offer potential quantum speedup on near-term quantum devices \cite{3842}.

Of the various suggested ways to merge classical machine learning techniques with quantum computing, the method introduced by Henderson et al. in \cite{3882} offers several advantages. It can be implemented on quantum circuits with fewer qubits and shallow gate depths, yet it can be applied to more practical use cases. This method employs quantum circuits as transformation layers to extract features for image classification using convolutional neural networks (CNNs). The transformation layers are referred to as quanvolutional layers, and the method is referred as a quanvolutional neural network (QuanvNN) in this research article.

A crucial query arose regarding whether the features generated by quanvolutional layers could enhance the classification accuracy of machine learning models. To investigate this, Henderson et al. have conducted a study where randomly generated quantum circuits were used to compare the classification accuracy of QuanvNN with a standard CNN. However, the findings did not demonstrate a clear advantage in classification accuracy over the classical model \cite{3882}. In a subsequent study \cite{3960}, QuanvNN was updated, implemented on quantum hardware (Rigetti’s Aspen-7-25Q-B quantum processing unit), and evaluated on a satellite imagery classification task. Nevertheless, the image classification accuracy of QuanvNN was not improved in comparison to that of a traditional CNN algorithm.

The work of Mari \cite{4013} provided an implementation of QuanvNN on a software quantum computing simulator called PennyLane \cite{3664}. Their approach differs from that of Henderson et al. in that the output of the quantum circuit, which is a set of expectation values, is directly fed into the subsequent neural network (NN) layer, whereas Henderson et al.\ \cite{3882} transformed it into a single scalar value using a classical method. The proposed method was tested on the MNIST dataset \cite{3937}, which consists of handwritten digits, using 50 training and 30 test images. However, no clear improvement in classification accuracy by QuanvNN over NN was shown in Mari’s study.

In our previous research \cite{4241}, we extended Mari’s QuanvNN by utilising a randomly generated quantum circuit with four qubits, 20 single axis rotations, and 10 controlled NOTs (CNOTs) to enhance image classification accuracy when compared to a classical fully connected NN. Specifically, the extended QuanvNN approach improved the accuracy of MNIST and CIFAR-10 datasets (photographic 10 class image dataset \cite{4063}) from 92.0\%\ to 93.0\%\ and from 30.5\%\ to 34.9\%, respectively \cite{4241}. We also proposed a new model, neural network with quantum entanglement (NNQE), that incorporates a strongly entangled quantum circuit with four qubits, 20 three axis rotations, 20 CNOTs, and Hadamard gates. This model further increased image classification accuracy against MNIST and CIFAR-10 to 93.8\%\ and 36.0\%, respectively \cite{4241}. However, using QuanvNN or NNQE was found to degrade the image classification accuracy when applied to a more complicated German Traffic Sign Recognition Benchmark (GTSRB) dataset (43 class real-life traffic sign images \cite{3847}) in comparison with the classical NN accuracy from 82.2\%\ to 71.9\%\ (QuanvNN) and to 73.4\%\ (NNQE) \cite{4241}.

The concept of using a quantum circuit as a pre-processing filter for image classification tasks has been extended by the introduction of quantum pre-processing filter (QPF) by the authors in \cite{4547}. In \cite{4547}, a much simplified quantum circuit, i.e.\ a four qubit quantum circuit with Y rotations for encoding and two CNOTs, was introduced. By applying the QPF approach, the results showed that the image classification accuracy based on MNIST and EMNIST (handwritten 47 class digits and letters \cite{3968}) datasets were improved against classical NN from 92.0\%\ to 95.0\%\ and from 68.9\%\ to 75.8\%, respectively. However, tests using the proposed QPF approach against GTSRB showed again a degradation in the classification accuracy from 81.4\%\ to 77.1\%\ \cite{4547}.

In this study, we first extend the application of QPF using two CNOTs from multi-class classification to binary classification against all possible different pairs of image classes. For 10 classes, e.g. MNIST, the total number of pairs is $10 \times 9 = 90$. For 43 classes, e.g. GTSRB, the total number of pairs is $43 \times 42 = 1,806$. The proposed method achieves a higher image classification accuracy of 98.9\%\ compared to 92.5\%\ against MNIST using NN. The image classification accuracy was further improved to 99.2\%\ by the application of QPF. While the image classification against GTSRB was improved from 81.4\%\ to 93.5\%\ using the proposed binary image classification method, the application of QPF degraded the image classification accuracy from 93.5\%\ to 92.0\%, similar to our previous results. We note that practical application of the proposed binary classification approach requires an additional categorisation method to extract training and testing images corresponding to the chosen classes from larger samples. This additional categorisation method is outside of the scope of the current study and is left for further study. In addition, we have applied the two CNOTs QPF to CIFAR-10 and EMNIST datasets and observed the binary image classification accuracy improvements from 71.2\%\ to 76.1\%\ and from 97.8\%\ to 98.3\%, respectively.

Secondly, we apply QPF to cases using a smaller number of training and testing samples, i.e. 80 training samples and 20 testing samples per class. The use of a smaller number of samples is considered in application where faster training and testing is required. In order to derive statistically stable results, we conducted the experiment with 100 trials choosing randomly different training and testing samples and averaged the results. The result showed that the application of QPF did not improve the image classification accuracy against MNIST and EMNIST but improved it against CIFAR-10 and GTSRB from 65.8\%\ to 67.2\%\ and from 90.5\%\ to 91.8\%, respectively. While the exact cause of this phenomenon is currently under investigation, this result is significant in understanding the effects of QPF in machine learning methods. In order to support our claims, we have made our source codes available at \verb+https://github.com/hajimesuzuki999/qpf-bic+.

The structure of this research paper is as follows: Section 2 outlines the methodology of our proposed model. Section 3 provides a detailed account of our experimental setup. Section 4 contains the results and discussion. Finally, in Section 5, we present our conclusions.

\section{Methodology}

\begin{figure}[h]
  \includegraphics[width=\linewidth]{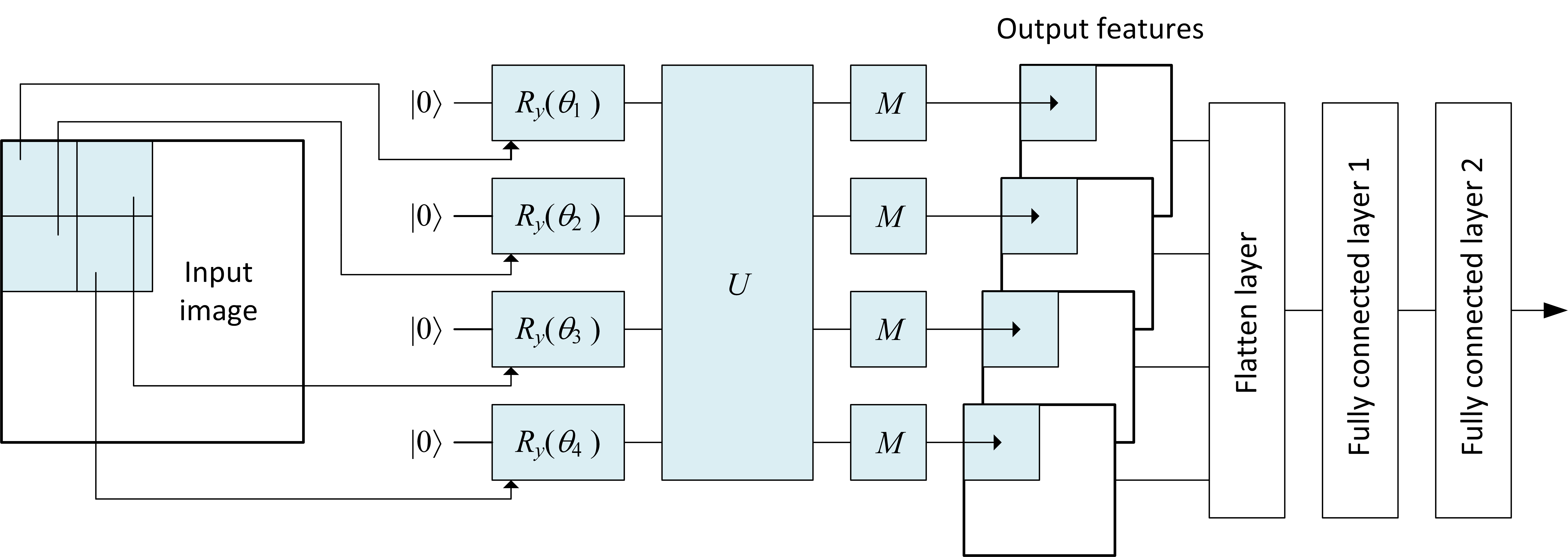}
  \caption{The architecture of the QPF model.}
  \label{fig:architecture}
\end{figure}

The architecture of QPF was first proposed in \cite{4547}. For the sake of completeness, we reproduce the description of QPF in this section. Figure~\ref{fig:architecture} shows the architecture of the proposed QPF. The method assumes that the input image is a two-dimensional matrix with size $m$-by-$m$, and the pixel value $x$, follows $0 \leq x \leq 1$. An extension to a multi-channel pixel image is considered as straightforward. A section of size $n$-by-$n$ is extracted from the input image. The proposed QPF uses $n = 2$. This $2 \times 2$ section of the input image is referred as QPF window.

\begin{figure}
\centering
  \includegraphics[scale=0.8]{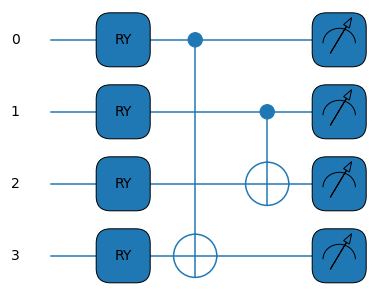}
  \caption{QPF with two CNOTs.}
  \label{fig:two_cnots}
\end{figure}

The outputs from the Y rotation gates are fed to the quantum circuit referred as $U$ in Figure~\ref{fig:architecture}. Measurements, referred as $M$ in Figure~\ref{fig:architecture}, are performed on the output of the quantum circuit $U$. The structure of the quantum circuit $U$ is further detailed in Figure~\ref{fig:two_cnots}. In \cite{4547}, we conducted experiments with different CNOTs arrangement (quantum entanglement property of quantum mechanics) and found that the arrangement as given in Figure~\ref{fig:two_cnots} showed superior improvements in image classification accuracy.

The outputs from the measurement operations are given as expectation values between $-1$ and $1$, and form output features. We note that the total number of parameters in the input image $( m \times m )$ is the same as the total number of parameters in the output features $( 4 \times ( m/2 ) \times ( m/2 ) )$. The output features are made into a one-dimensional vector by the fatten layer. The number of nodes of the output of the flattening layer is $m \times m$. The nodes are fully connected by the first fully connected layer 1. The output of the fully connected layer 2 has the number of nodes equal to the number of classes.

\section{Experiment}

The method proposed has been implemented using MATLAB and Python. The Adam optimiser and a batch size of 128 have been used for training the network. Four datasets were utilised: MNIST, EMNIST, CIFAR-10 and GTSRB.

\begin{figure}[h]
  \includegraphics[width=\linewidth]{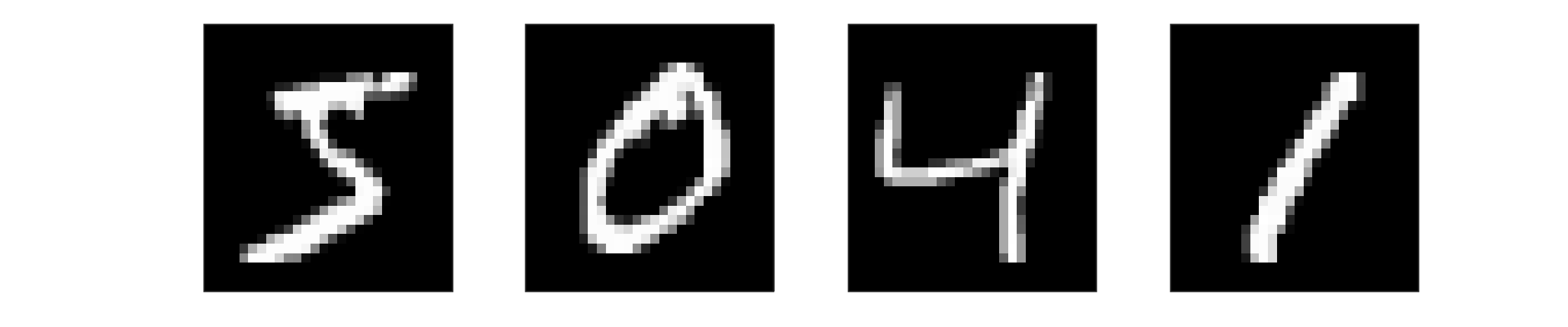}
  \caption{Example MNIST dataset images.}
  \label{fig:mnist}
\end{figure}

The MNIST dataset comprises of 60,000 training and 10,000 testing images of handwritten digits ranging from 0 to 9 \cite{3937}. Each image is of size 28 by 28 pixels. The original images are represented in grayscale with pixel values between 0 and 255, which are normalised by dividing them by 255. Figure~\ref{fig:mnist} shows some examples of images from the MNIST dataset.

The EMNIST dataset comprises 112,800 training and 18,800 test images of handwritten digits and letters making up 47 classes \cite{3968}. The image size and scaling are the same as MNIST dataset. The CIFAR-10 dataset comprises 50,000 training and 10,000 test images of 10 class photographic images \cite{4063}. The original images are in RGB color, which were converted into grayscale between 0 and 255 and then scaled by dividing them by 255.

\begin{figure}[h]
  \includegraphics[width=\linewidth]{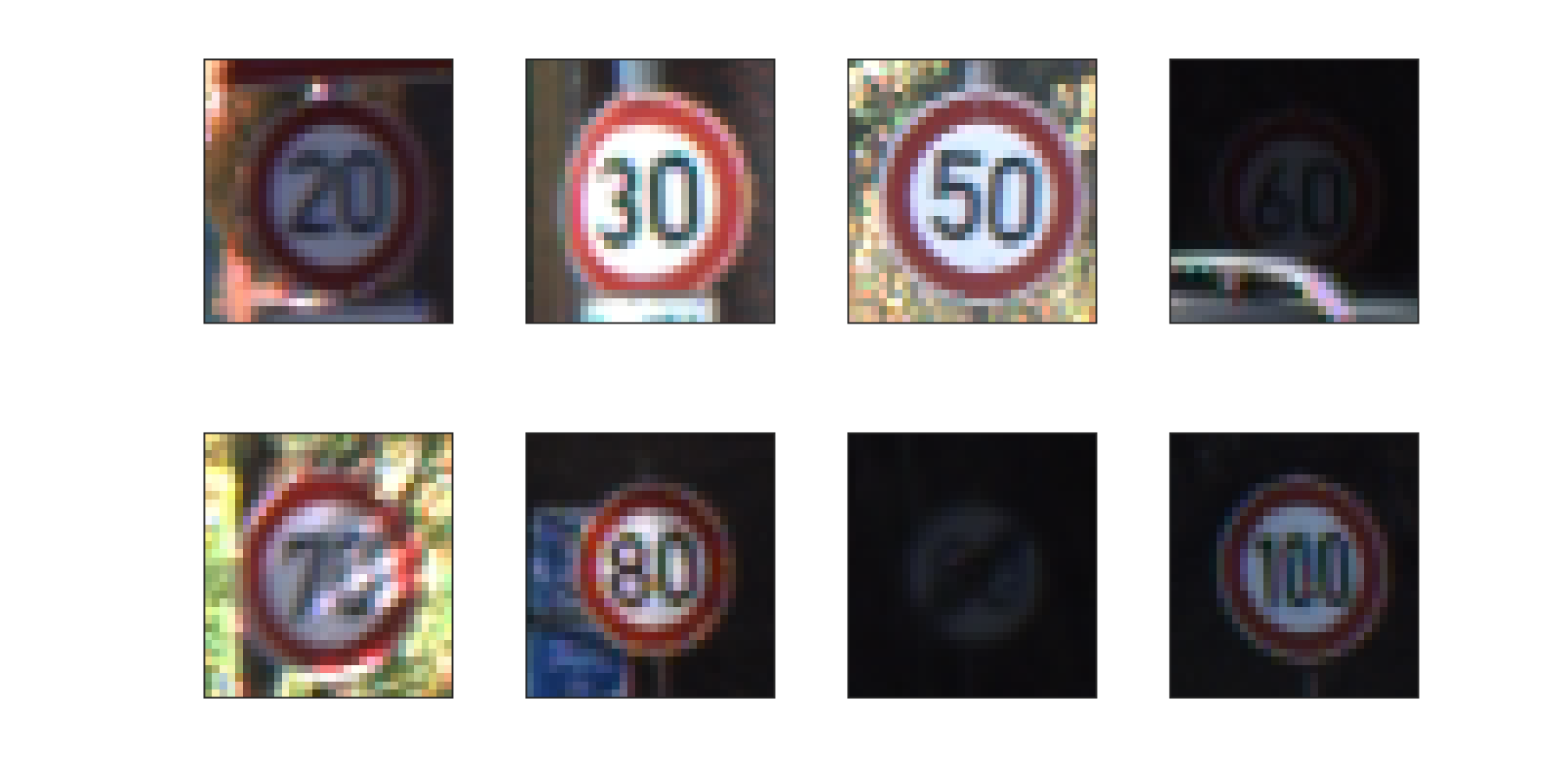}
  \caption{Example GTSRB dataset images}
  \label{fig:gtsrb}
\end{figure}

The GTSRB dataset \cite{3847} comprises 34,799 training and 12,630 test images of 43 different classes of traffic signs. These images are actual pictures of traffic signs captured under different conditions. The size of the original images varies between $15 \times 15$ and $222 \times 193$ pixels. However, in this experiment, all images have been scaled to a size of $32 \times 32$ pixels. The images in the dataset are initially in RGB colour format, but they were converted into grayscale, with pixel values ranging between 0 and 255. Then, the pixel values were scaled down by dividing them by 255 to normalise the data. Figure~\ref{fig:gtsrb} provides some examples of images from the GTSRB dataset.

Table~\ref{tbl:parameters} summarises the parameters of the three image datasets used in the experiment.

\begin{table}
\centering
\begin{tabular}{c|cccc}
& {\bf MNIST} & {\bf EMNIST} & {\bf CIFAR-10} & {\bf GTSRB} \\ \hline
{\bf Image size} & $28 \times 28$ & $28 \times 28$ & $32 \times 32$ & $32 \times 32$ \\
{\bf Number of colour channel} & 1 & 1 & 3 & 3 \\
{\bf Number of classes} & 10 & 47 & 10 & 43 \\
{\bf Number of class pairs} & 90 & 2,162 & 90 & 1,806 \\ \hline
\end{tabular}
\caption{Parameters of image datasets used in the experiment.}
\label{tbl:parameters}
\end{table}

\section{Results and Discussion}

\begin{figure}
\centering
\subfloat[Using NN.]{\includegraphics[scale=0.6]{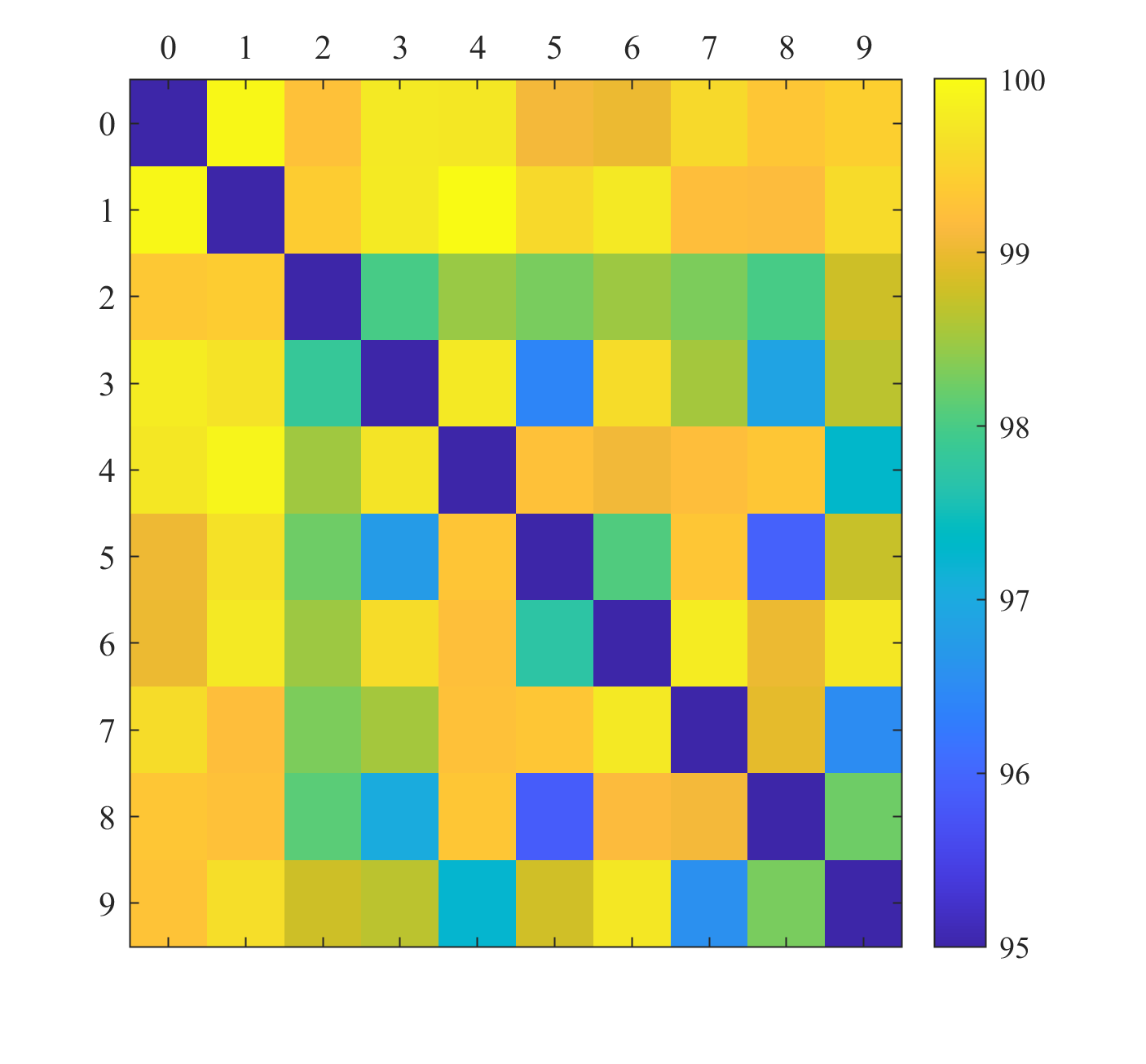}}\\
\subfloat[Using QPF-NN.]{\includegraphics[scale=0.6]{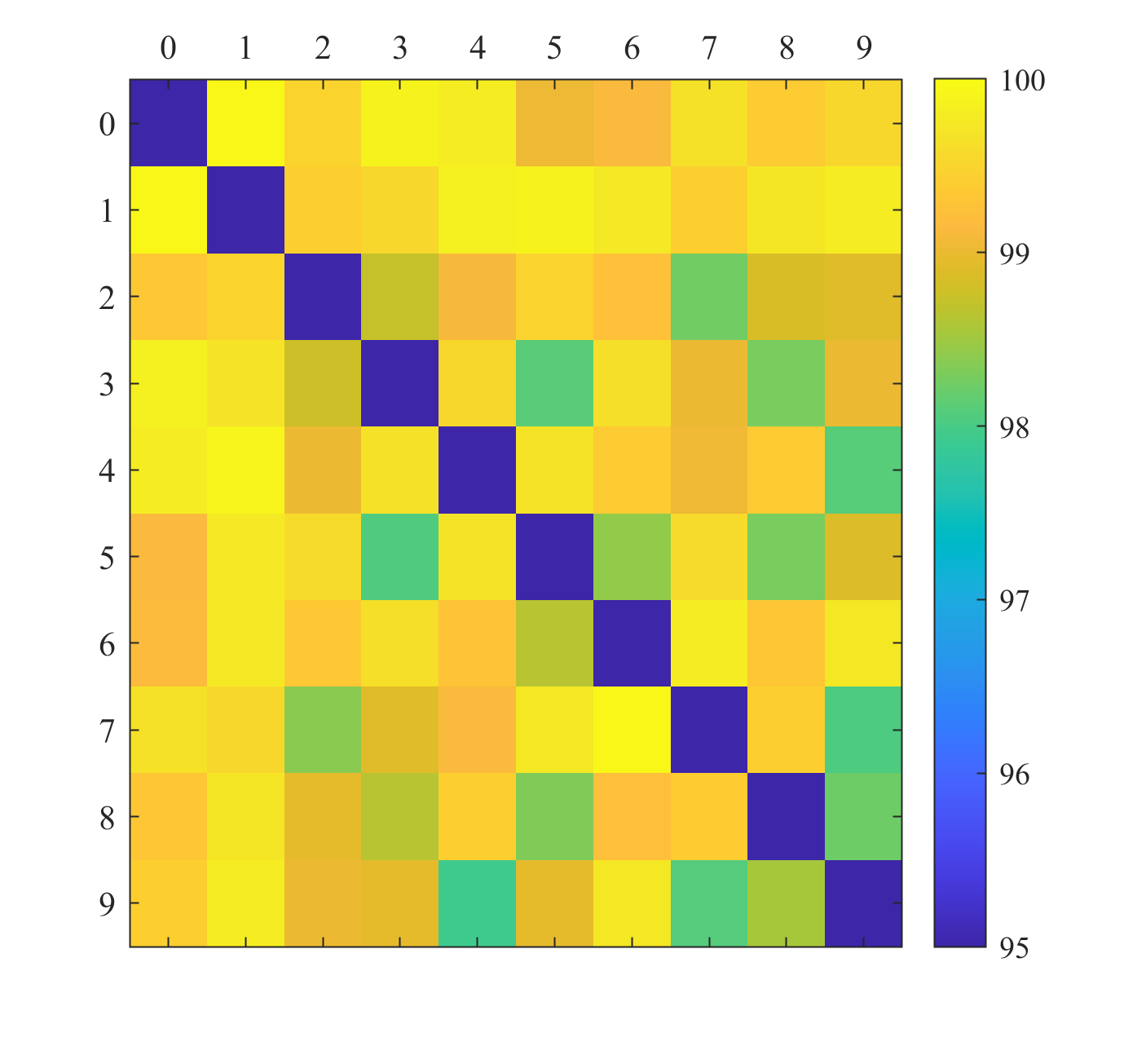}}
\caption{Testing accuracy against MNIST using all data.}
\label{fig:testing_accuracy_mnist}
\end{figure}

First, we use all available training and testing samples to perform binary image classification against all different pairs of classes using NN. The results are shown in Figure~\ref{fig:testing_accuracy_mnist} (a) for MNIST dataset. In this graph, the testing accuracy for the given pair is shown by a different colour. For example, classifying the number 0 against 1 achieves close to 100\%\ accuracy, as shown in light yellow. In comparison, we can observe that the testing accuracy for number 5 against 8 is poor, about 96\%\ accuracy. This is due to the similarity in the shapes of the handwritten numbers 5 and 8. Additionally, other class pairs such as 3 and 5, 3 and 8, 4 and 9, and 7 and 9 also have similar shapes, leading to lower testing accuracy for those pairs. On average, the binary image classification using classical NN against MNIST achieved 98.9\%\ testing accuracy using all data. Figure~\ref{fig:testing_accuracy_mnist} (b) shows the results for QPF-NN against MNIST using all data. A similar result is obtained with an improvement in average image classification accuracy of 99.2\%. When it is applied to EMNIST and CIFAR-10, QPF-NN improved the testing accuracy over NN from 97.8\%\ to 98.3\%\ and from 71.2\%\ to 76.1\%, respectively.

\begin{figure}
\centering
\subfloat[Using NN.]{\includegraphics[scale=0.6]{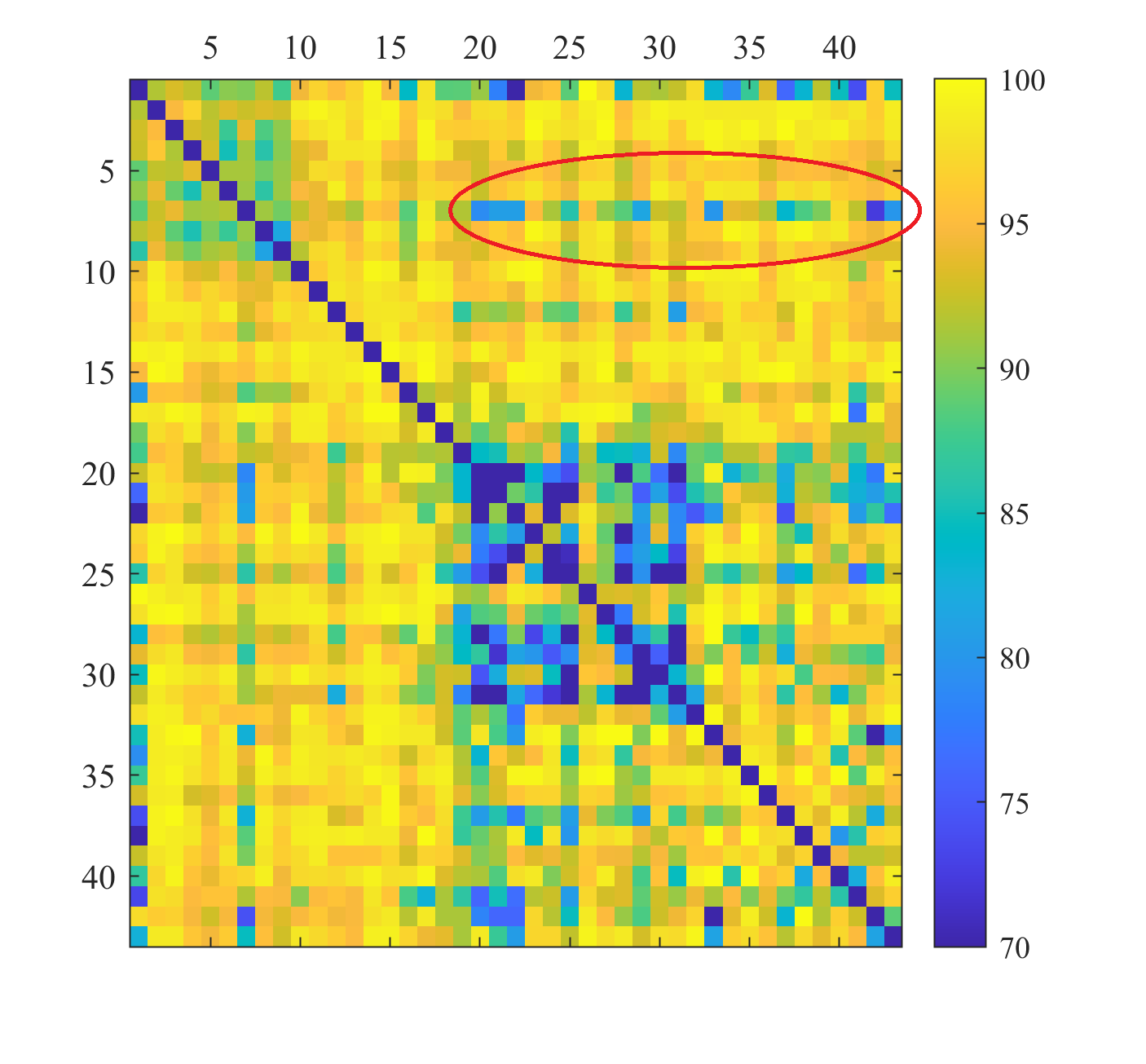}}\\
\subfloat[Using QPF-NN.]{\includegraphics[scale=0.6]{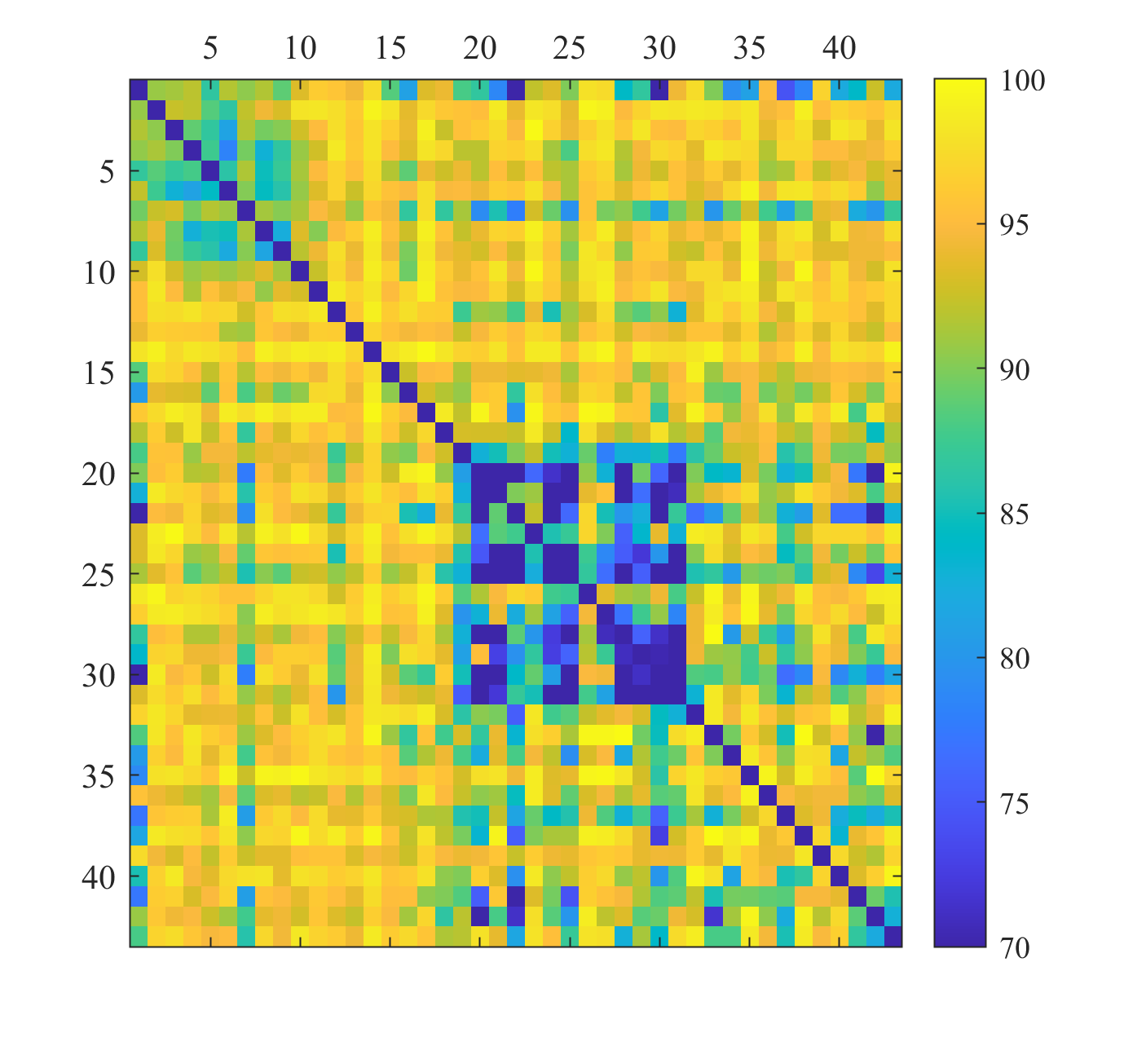}}
\caption{Testing accuracy against GTSRB using all data.}
\label{fig:testing_accuracy_gtsrb}
\end{figure}

Figure~\ref{fig:testing_accuracy_gtsrb} (a) shows the NN results against GTSRB using all data. We can identify pairs of classes that produce high testing accuracy and those that do not. For example, class 7 shows lower testing accuracy against many of the classes between 20 to 43, compared to class 6 or 8, as indicated by the red oval.  Referring to \cite{3847}, Fig.~1, class 7 corresponds to 80~km/h sign with a diagonal strip. This class also has a smaller number of samples (Approx.\ 1/3) compared to class 6 or 8. Detailed examination of this graph may provide further insights, however, we focus on the effects of the application of QPF, and hence this is left for a future study.  The average testing accuracy over all different pairs is 93.5\%. In comparison, Figure~\ref{fig:testing_accuracy_gtsrb} (b) shows the QPF-NN results against GTSRB using all data. Similar results were obtained with a reduced average testing accuracy of 92.0\%.

\begin{figure}[h]
  \includegraphics[width=\linewidth]{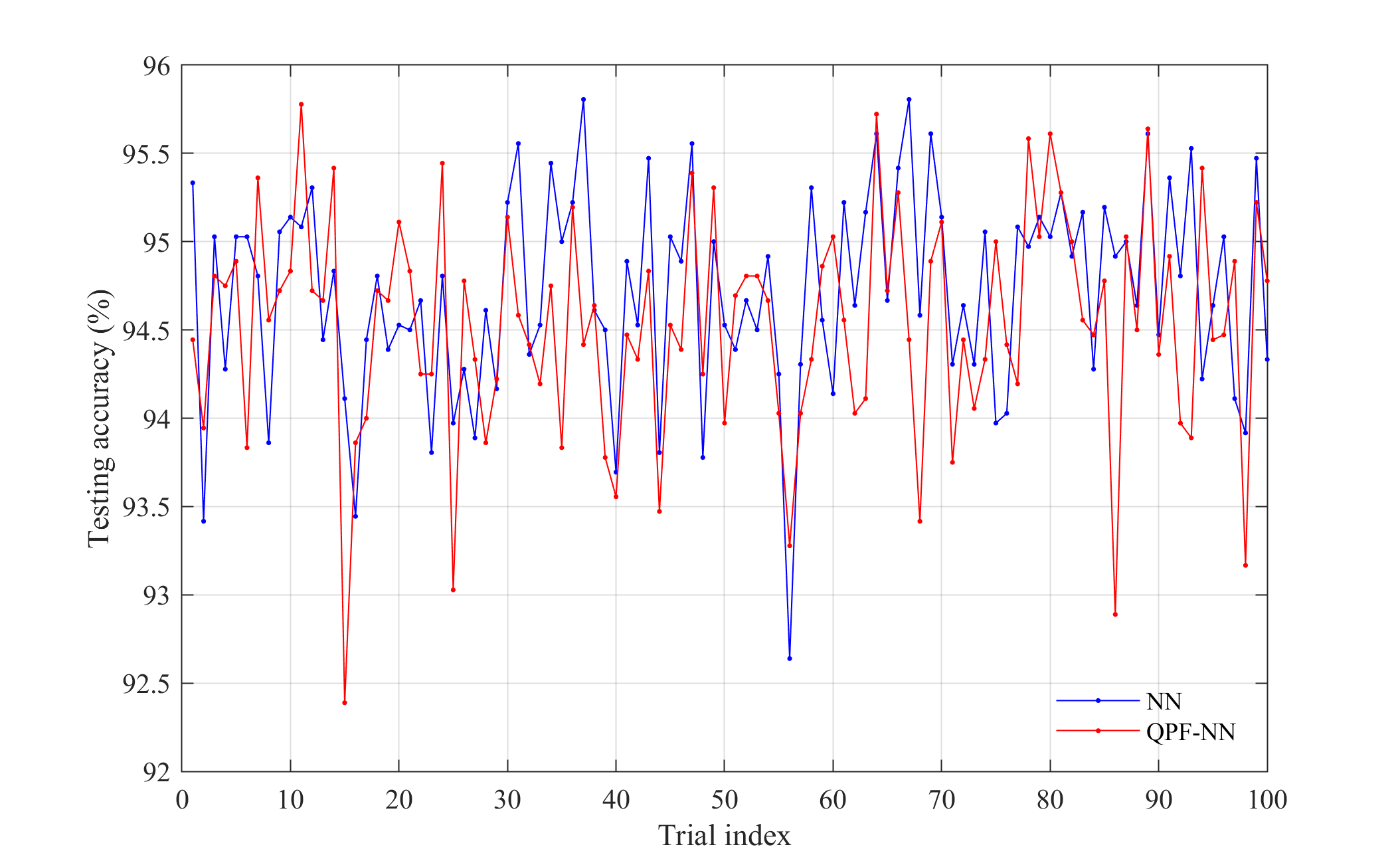}
  \caption{Testing accuracy against MNIST using 100 samples and 100 trials.}
  \label{fig:mnist_100}
\end{figure}

Secondly, we performed 100 trials with each trial extracting 80 training samples and 20 testing samples per class randomly to perform training and testing. Figure~\ref{fig:mnist_100} shows the variation of testing accuracy as a function of a trial index when NN and QPF-NN are used against MNIST. We observe that variation is relatively large (approximately 3\%) which shows the importance of performing multiple trials and averaging the results to obtain statistically stable results. On average, the testing accuracy of 94.7\%\ and 94.5\%\ was obtained for NN and QPF-NN, respectively. In this case, the application of QPF shows minimal effects. Similarly, we observed minimal effects of using QPF-NN over NN against EMNIST having the same testing accuracy of 94.0\%.

\begin{figure}[h]
  \includegraphics[width=\linewidth]{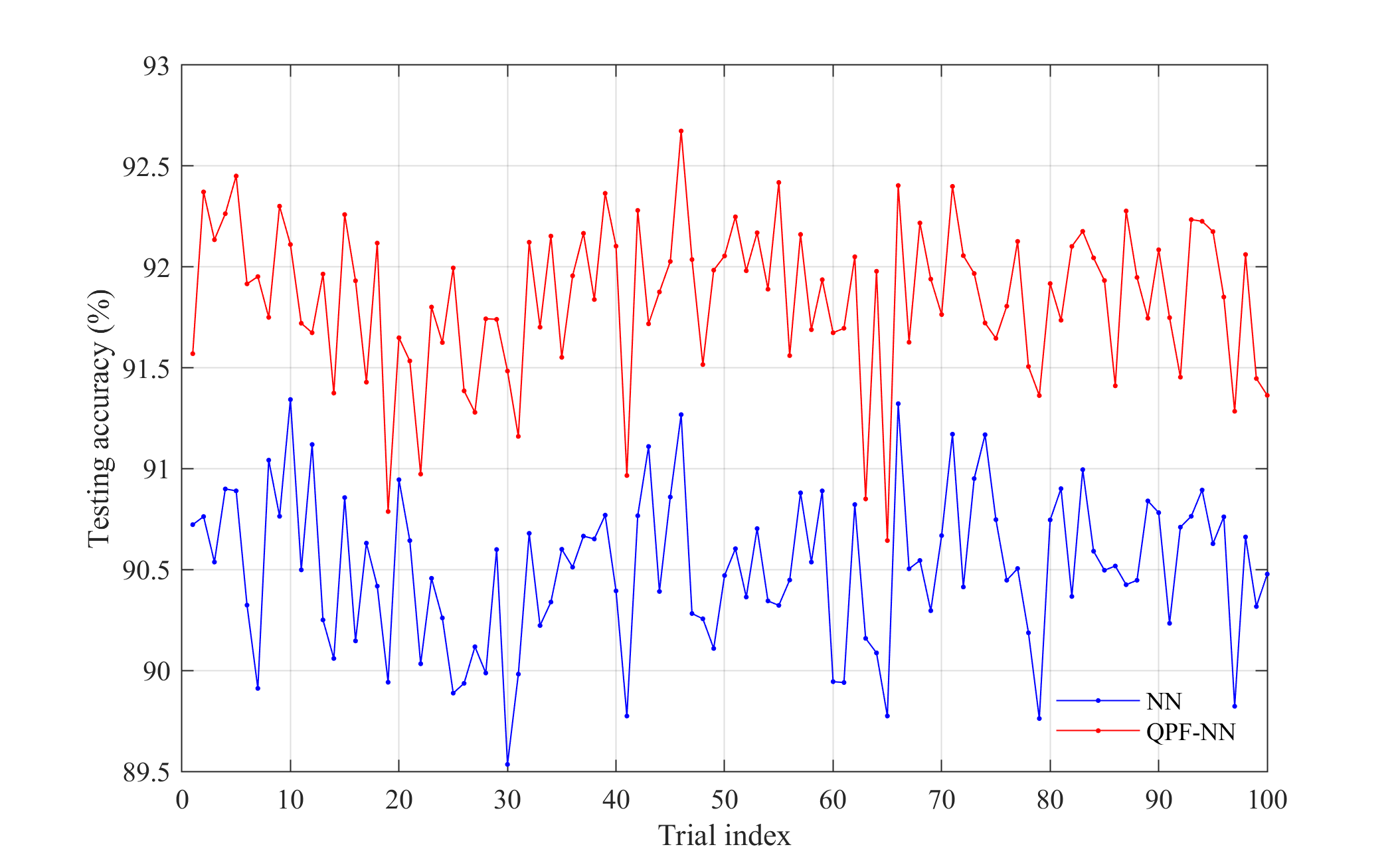}
  \caption{Testing accuracy against GTSRB using 100 samples and 100 trials.}
  \label{fig:gtsrb_100}
\end{figure}

Figure~\ref{fig:gtsrb_100} shows the results against GTSRB. We observe that the variation is relatively small (approximately 1\%) which may be due to a larger number of class pairs (1,806 for GTSRB compared to 90 for MNIST) over which the testing accuracy is averaged for each trial. Importantly, the application of QPF shows improvement over NN against GTSRB, which was not observed in any of our previous experiments. It is also notable that QPF-NN always improved the testing accuracy over NN in any of the 100 trials. We note that the same set of training and testing samples was used for NN and QPF-NN for each trial. We have observed a similar result with CIFAR-10 with an improved test accuracy from 65.8\%\ to 67.2\%. A summary of the testing accuracy results is shown in Table~\ref{tbl:summary}.

\begin{table}
\centering
\begin{tabular}{c|cccc}
& {\bf MNIST} & {\bf EMNIST} & {\bf CIFAR-10} & {\bf GTSRB} \\ \hline
{\bf All data, NN} & 98.9\% & 97.8\% & 71.2\% & 93.5\% \\
{\bf All data, QPF-NN} & 99.2\% & 98.3\% & 76.1\% & 92.0\% \\
{\bf 100 samples, NN} & 94.7\% & 94.0\% & 65.8\% & 90.5\% \\
{\bf 100 samples, QPF-NN} & 94.5\% & 94.0\% & 67.2\% & 91.8\% \\ \hline
\end{tabular}
\caption{A summary of testing accuracy results.}
\label{tbl:summary}
\end{table}

\section{Conclusion}

This study aimed to evaluate the performance of a proposed binary image classification method using a QPF model with 4 qubits and 2 CNOTs. In our previous research we have shown that QPF is used for efficient image feature extraction while existing quantum circuits demand high computation and multiple layers to extract image features. Similar to the previously reported multi-class classification case, the proposed QPF model improved binary image classification accuracy against MNIST, EMNIST, and CIFAR-10 but we observed a slight decrease in the performance against GTSRB using all training and testing samples. However, when applied to the cases with a smaller number of training and testing samples, QPF improved image classification performance against CIFAR-10 and GTSRB, which shows better generalisation of our QPF model for smaller number of samples compared to previous classical NN models, which mostly requires a larger number of sample to generalise \cite{3889}. The results presented in this article provide further insights into the effects QPF on machine learning algorithms. Further research will be conducted as part of future work to investigate the potential of QPF to assess the scalability of the proposed approach to larger and complex datasets.

\section{Acknowledgment}

This research has been supported by Australian government Research Training Program and Commonwealth Scientific Industrial and Research Organization.

\bibliographystyle{unsrt}
\bibliography{hajimesbib}

\end{document}